\documentclass[letterpaper, 10 pt, conference]{ieeeconf}
\usepackage{subfig}
\usepackage{graphicx} 
\usepackage{epsfig} 
\usepackage{amsmath} 
\usepackage{amssymb}  
\usepackage{setspace}
\usepackage{booktabs}
\usepackage{hyperref}
\usepackage{bm}
\usepackage{caption}      
\usepackage{xspace}
\usepackage{url}
\usepackage{wrapfig}
\usepackage{mathrsfs}
\usepackage{array}
\usepackage{euscript}
\usepackage{hybridsystem}
\usepackage{color}
\usepackage{balance}
\usepackage{tikz}
\IEEEoverridecommandlockouts

\title{Deep Visual Perception for Dynamic Walking on Discrete Terrain}
\author{Avinash Siravuru,  Allan Wang, Quan Nguyen, and Koushil Sreenath %
		\thanks{This work is supported in part by grants from the Pennsylvania Infrastructure Technology Alliance, the Google Faculty Research Award, and in part by NSF grant IIS-1526515.}%
		\thanks{A. Siravuru, A. Wang, and Q. Nguyen are with the Department of Mechanical Engineering, Carnegie Mellon University, Pittsburgh, PA 15213, USA
			{\tt\small \{avinashs,allanwan,quannguyen\}@cmu.edu}.}%
		\thanks{K. Sreenath is with the Dept. of Mechanical Engineering, University of California, Berkeley, CA 94720, {\tt\small koushils@berkeley.edu}.}%
} 
            
\date{}
\begin{document}

\maketitle

\begin{abstract}
Dynamic bipedal walking on discrete terrain, like stepping stones, is a challenging problem requiring feedback controllers to enforce safety-critical constraints. To enforce such constraints in real-world experiments, fast and accurate perception for foothold detection and estimation is needed. In this work, a deep visual perception model is designed to accurately estimate step length of the next step, which serves as input to the feedback controller to enable vision-in-the-loop dynamic walking on discrete terrain.  In particular, a custom convolutional neural network architecture is designed and trained to predict step length to the next foothold using a sampled image preview of the upcoming terrain at foot impact. 
%
The visual input is offered only at the beginning of each step and is shown to be sufficient for the job of dynamically stepping onto discrete footholds. Through extensive numerical studies, we show that the robot is able to successfully autonomously walk for over $100$ steps without failure on a discrete terrain with footholds randomly positioned within a step length range of $[45 : 85]$ centimeters. 
\end{abstract}
\section{Introduction}
\label{ch:intro}
\subsection{Problem Definition} The objective of this work is to build and systematically evaluate a Deep Visual Perception system to be used by a planar dynamic walking robot in order to autonomously walk on discrete terrain.
We build a realistic visual simulator to generate the robot's first person view and combine it with an accurate physics simulator of the bipedal robot walking on discrete terrain. The physics simulator also contains an inner-loop safety-critical controller that can generate stable and safe limit cycle walking of a desired step length \cite{RSS2017_DiscreteTerrain_Walking}. In this setup, we train a deep neural network to estimate the step length (distance to the next stepping location) using a single sampled image preview that is obtained at the beginning of each step. Detecting footholds and estimating distance is a classic object localization problem similar to object grasping in robotic manipulation, however in the case of locomotion there are additional challenges due to the time-critical and safety-critical nature of the problem.

Note that,  we limit our attention to only autonomous planar walking, and accordingly, only predict step length information. This simplification allows us to keep the focus on the visual simulator development, custom CNN design (to bound worst-case estimate) and perception-control integration. However, the method itself can be extended to 3D walking without loss of generality as evidenced by (a) our prior work on dynamic walking over terrain with varying step width and step height in addition to step length \cite{CDC2016_3DWalking_SteppingStones,RSS2017_DiscreteTerrain_Walking}, and (b) the perception system in this paper that takes an image rendered from a 3D scene as input without making any geometric simplifications due to the planar walking.  

\subsection{Motivation} 
\begin{figure}
\centering			             
\includegraphics[width=0.8\columnwidth]{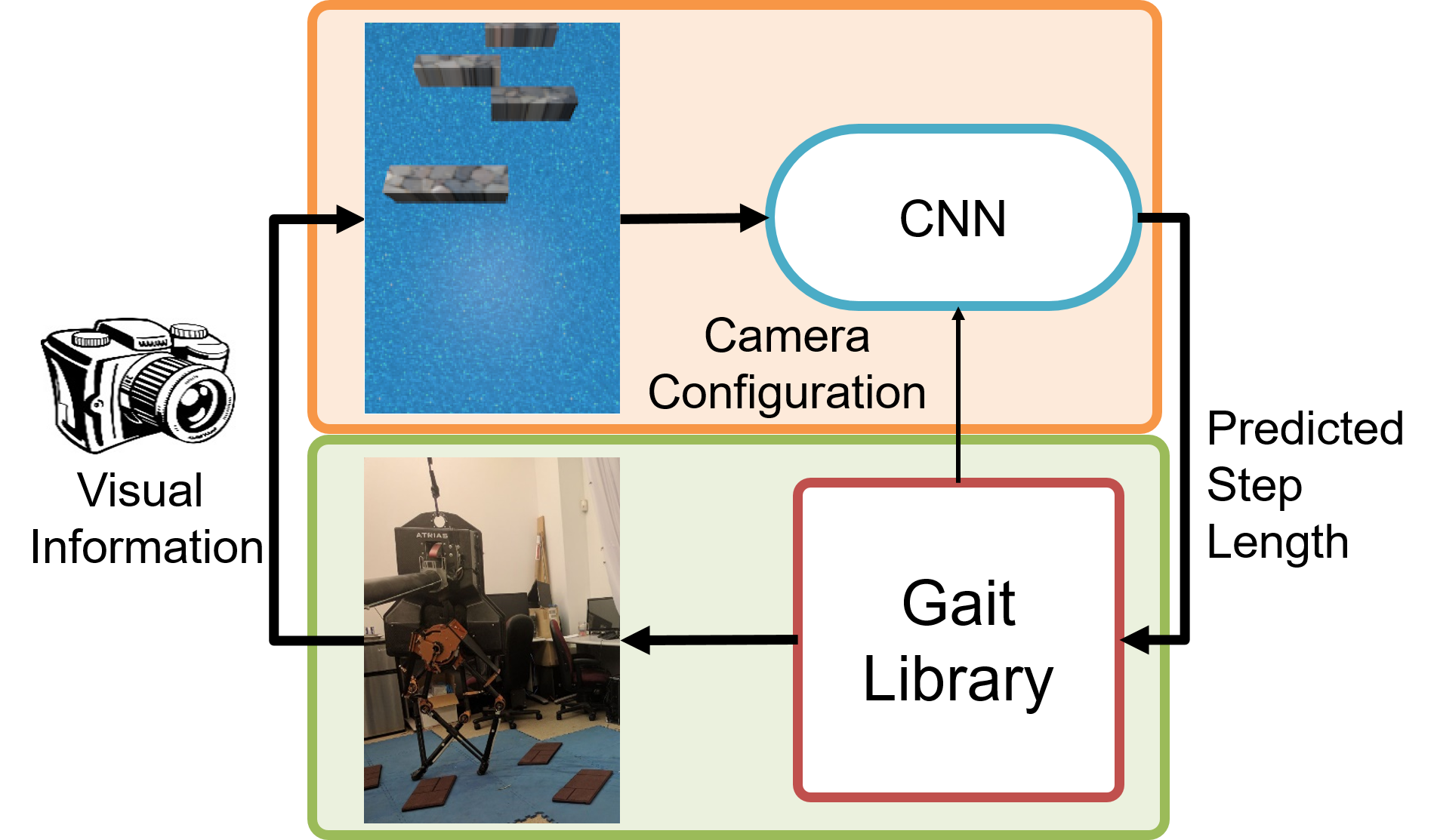}
\label{fig:papersummary}
\caption{At the beginning of each step, the convolutional neural network predicts the step length based on per-step visual preview from a torso-mounted camera. A suitable limit cycle is chosen from the gait library \cite{RSS2017_DiscreteTerrain_Walking} and executed to achieve the predicted step length.}
\vspace{-15pt}
\end{figure}

Conventionally, robot perception involves parsing the entire scene, labeling objects of interest, and feeding this information for planning and control. 
As the number of tasks increase and the decision-making time drops, searching the entire scene for all tasks is expensive and trade-offs are inevitable. Reactive vision-in-the-loop control for sub-tasks, wherever possible, reduces overhead on higher-level planners and injects more dynamism into the system. The objective of this work is to serve this need for the sub-task of walking safely on discrete terrain.

In fact, experiments with humans and cats have shown that during over $50-60 \%$ time of the gait cycle, gaze is invested in target detection and the higher-level spatial navigation problem \cite{patla1997understanding}. When required to walk on complex terrain requiring accurate foot placement, humans operate with intermittent visual samples of the foothold location and use the information in a feed-forward manner to adjust step length just $1-2$ steps a priori \cite{patla1997understanding, matthis2013rsb}. For walking on discrete terrain, the key finding here is that, instead of active modulation of gait through visual feedback during the entire stance phase, humans prefer to adjust their gait one-two step ahead using an intermittent visual preview and execute an energy-efficient ballistic motion \cite{matthis2017critical}. If the foothold remains constant, continuous visual feedback may even be unnecessary. Our work is inspired by these biological findings and we wish to elicit similar behavior from the dynamic bipedal robot ATRIAS. Our paper makes the following contributions:

\begin{enumerate} 
\item We present a custom deep convolutional neural network (CNN) architecture for the task of estimating next step location given a sampled visual preview of the terrain at the end of the current step. 
A synthetic outdoor dataset with dynamic real world textures, varying lighting conditions and camera positions is also developed. The network, trained on this dataset, returns an average prediction error of $1.62$ cm and a worst-case error of $10.38$ cm. 
\item We integrate the CNN-based step length estimator with an inner-loop safety-critical controller to enable vision-in-the-loop dynamic walking on discrete terrain. 
The robot with the visual step-length estimator and the safety-critical controller is shown to successfully walk at least 100 steps without failure, with the step lengths randomly sampled from a uniform distribution of $45-85$ cm. 
\end{enumerate}
    
\section{Related Work}
\textbf{Deep Learning for Robotic Vision:}
Recent advances of Deep Learning in fields like computer vision, natural language processing, speech recognition, etc. has lead to tremendous research interest in extending these gains to robotics. Large amounts of data available for network training and parallel computation helped accelerate this effort. From learning end-to-end visuomotor policies for object manipulation \cite{levine2016end}, to learning to fly UAVs in cluttered environments \cite{sadeghi2016cadrl}, or in self-driving cars \cite{nvidiacar,chen2015deepdriving}, deep learning is impacting all major robotics sub-domains. In humanoid robotics also, CNNs were used for innovative applications like surface friction estimation from images, to help in slip prediction \cite{matcnn}. However, it is very challenging to build end-to-end fully data-driven policies for stable and safe limit cycle walking, let alone on discrete terrain. 

\textbf{Perception in Legged Locomotion:}
Perception for bipedal locomotion primarily focused on foot-step planning for statically stable or linear dynamical model-based walkers. Usually, LIDAR-camera combination is preferred in this case. Accurate high resolution depth data obtained from Lidar is used for safe footstep detection and planning \cite{chestnutt2009biped,Fallon2015}. Unlike these walkers, dynamics robots have point feet, move much faster and therefore need faster execution and the ability to pick footholds of any size. This makes the search problem harder on the full 3D map. 
Vision-in-the-loop walking with gait adjustment (comparable to our approach) was implemented on a Quadruped in \cite{bajracharya2013high} to operate on steep slopes and dense vegetation. However, the problem of discrete terrain is not addressed. We believe our solution is complimentary to their effort and a combined solution could pave the way for true rough terrain navigation. 



\section{Robot Model and Controller Summary}
\label{sec:modelandcontrol}
Having presented an overview of related work, we will now briefly develop the dynamical model and controller for achieving stable walking with precise foot placements. 

\subsection{Dynamical Model for Walking}
\label{subsec:model}
We consider the ATRIAS bipedal robot with configuration $q$ and state $x = (q,\dot q)$ that denotes the generalized positions and velocities of the robot, with $u$ denoting the joint torques. A hybrid model of walking can then be expressed as
\begin{align}
\label{eq:hybridsystem}
\left\{\begin{array}{llll}\dot{x} & =& f(x)+g(x)u  & x \notin {\cal S}
\\ x^+ &=& \Delta(x^-) & x \in {\cal S}, \end{array}\right.
\end{align}
where ${\cal S}$ is the switching guard, $\Delta$ is the reset or impact map, and the superscripts $-$ and $+$ denoting pre- and post-impact variables respectively. A detailed description of the robot and a derivation of its model can be found in \cite{RAHUAKGR14}.


\subsection{Periodic Gait Design using Virtual Constraints}
\label{subsec:gait_library}
Virtual constraints are kinematic relations that synchronize the evolution of the robot's coordinates via continuous-time feedback control, thereby simplifying control of high degree-of-freedom systems, \cite{WGCCM07}. Virtual constraints are expressed as an output vector
\begin{align}
	\label{eq:yPhase}
	y(q) = h_0(q) - h_d(s(q),\alpha),
\end{align}
to be asymptotically zeroed by a feedback controller, with one virtual constraint typically imposed per each actuator. Here $h_0(q)$ specifies the variables to be controlled, and $h_d(s,\alpha)$ specifies the desired evolution of the controlled variables, parametrized by the the coefficient $\alpha$ and the gait phasing variable $s$ which goes from zero at the start of the gait to one at the end of the gait.


A nonlinear constrained optimization is used to find the coefficient $\alpha$ so as to create a periodic orbit satisfying a desired step length, while respecting physical constraints on torque, motor velocity, and friction cone. 
The cost function is taken as the integral of squared torques normalized by step length,
\begin{align}
	\label{eq:cost}
	J= \frac{1}{L_{step}}\int_{0}^{T} ||u(t)||^2_2~ dt,
\end{align}
and the constraints for the optimization are formulated as given in Table~\ref{tab:physicalConstraint}, see \cite[Sec.~6.6.2]{WGCCM07} for more details. The optimization is solved through a fast direct collocation framework from \cite{Jo2014}.

\begin{table}
	\centering
	\caption {Optimization constraints.}
	\label{tab:physicalConstraint}
	{\renewcommand{\arraystretch}{1.0}%
		\begin{tabular}{c l}
			Motor Toque         & $|u|\le 7$~Nm   \\
			Impact Impulse       & $F_e\le 15$~Ns  \\
			Friction Cone        & $\mu\le0.6$    \\
			Vertical Ground Reaction Force &
			$F_{st}^v\ge200$~N  \\
			Mid-step Swing Foot Clearance  & $h_f|_{s=0.5}\ge0.1$~m
		\end{tabular} }
        \vspace{-15pt}
	\end{table}

\begin{figure}
	\centering
    \subfloat[]{\centering
	\resizebox{0.42\columnwidth}{!}{\includegraphics{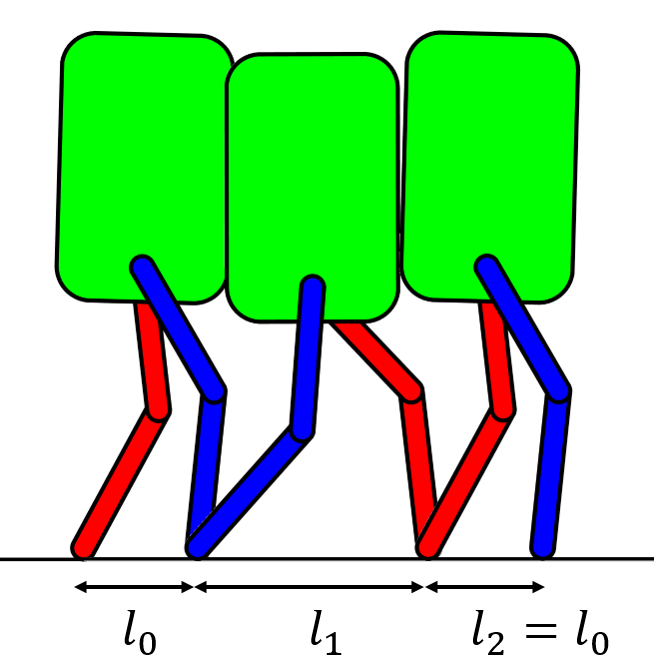}}
    \label{fig:TwoStepLengthOptimization}} 
    \subfloat[]{\centering
	\resizebox{0.42\columnwidth}{!}{\includegraphics{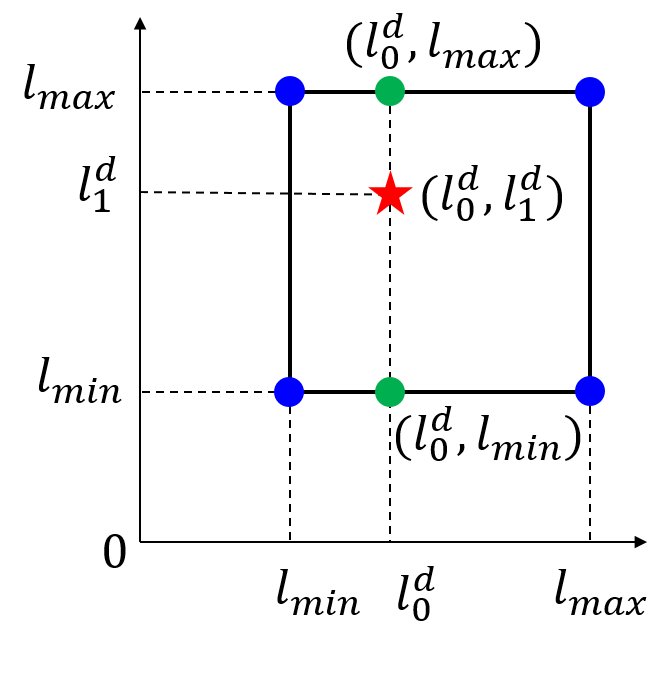}}
    \label{fig:GaitInterpolation}} 
	\caption{(a) 2-Step periodic walking with changing step lengths only. The walking gait is 2-step periodic therefore the step length of the second step and that of the initial condition are the same ($l_2=l_0$) and (b) Gait interpolation for the problem of changing step length.}
	\label{fig:TwoStepOptAndGaitInterp }
    \vspace{-15pt}
\end{figure}

Since walking over stepping stones involves changes in the step length, one way to easily transition between different step lengths is to create a 2-step periodic gait that explicitly considers the step length of the current step as well as the subsequent step.  The optimization process presented above can be used to design a 2-step periodic orbit with step lengths ($l_0$, $l_1$) with coefficient $\alpha(l_0, l_1)$ as shown in Fig.~\ref{fig:TwoStepLengthOptimization}.  

With a collection of 2-step periodic orbits, We can then transition at each step between multiple 2-step periodic orbits in a MPC-like manner with small transients.  However, to prevent an explosion of the number of periodic gaits that need to be optimized for, here we use only four 2-step periodic gaits corresponding to step lengths $(l_{min},l_{min}), (l_{min},l_{max}), (l_{max},l_{min}), (l_{max},l_{max})$ and use bilinear interpolation to find the coefficients $\alpha$ for a desired gait of step lengths $(l_0^d, l_1^d)$ as shown in Fig.~\ref{fig:GaitInterpolation}.
This work builds off recent work on periodic walking gait libraries in \cite{RSS2017_DiscreteTerrain_Walking, Quan_MarloSteppingStone, DaHaHaGrGr16, DaHaGr2017}.



\subsection{Feedback Control}
\label{subsubsec:IOlinearization}
The presented optimization results in a desired walking gait encoded through $h_d(s(q),\alpha)$ in \eqref{eq:yPhase}. The goal for the feedback control is to drive $y(q) \to 0$. In this work, we use a nonlinear feedback controller to enforce exponential stability for the hybrid system through input-output linearization, \cite{Ames2014Rapidly}. If $y(q)$ has vector relative degree 2, then its second derivative can be expressed through Lie derivatives as,
\begin{equation}
\label{yddot}
\ddot{y} = L^{2}_{f}y(\q,\dq) + L_{g}L_{f} y(\q,\dq) ~ u.
\end{equation}
We can then apply the following pre-control law
\begin{align}
\label{expControl}
u(q,\dot{q}) = u^{*}(q,\dot{q}) + (L_{g}L_{f} y(q,\dot{q}))^{-1} ~ \mu,
\end{align}
where the nominal feedforward component
$u^{*}(q,\dot{q}) := -(L_{g}L_{f} y(q,\dot{q}))^{-1} L^2_{f} y(q,\dot{q})$
with feedback $\mu=-K_p y-K_d \dot y$ stabilizes the system.
This combination of the 2-step periodic gait libary and the feedback controller results in precise footplacements with specified step lengths in a safety-critical manner.  Formal guarantees on safety for dynamic walking on discrete terrain can also be achieved through control barrier functions  \cite{Quan_MarloSteppingStone}.

\section{Direct Perception for ATRIAS}
\label{sec:perception}
Having presented the robot's dynamical model and the safety-critical controller designed to walk on discrete terrain when provided with accurate step length information, we will now present a systematic way to build and train a deep visual perception model that estimates the step length from a single monocular image. The system will take an input of the upcoming terrain through a front-facing camera at the beginning of each step.  This image is to be fed to a CNN to estimate the step length for the next step.  The step length estimate is 
then fed to the gait-library based controller to enable the robot to precisely land on the next foothold.

This CNN-based deep perception model has two critical components. Firstly, we need a large corpus of step-length annotated imagery of the robot's front person view while walking. This dataset is used to train the model for accurate step length estimation. The systematic methodology used to create this synthetic dataset is  described in the next subsection. Secondly, we need a suitable deep neural network architecture that can best approximate the complex non-linear mapping from image to step length estimate. The network needs to be tuned methodically to not only obtain the best test accuracy but also bound the worst-case prediction. These details are summarized after the next subsection.

\subsection{Dataset Generation}
To generate the image dataset, we use a popular open-source graphics software called Blender \cite{blender} to programatically generate realistic scenes. To create a discrete terrain scene, we need four key details: 1) Camera location and intrinsic parameters 2) Stepping stone location, 3)  Lighting model and location, and 4) Color and texture information of both the stepping stone and background. These parameters will be randomized in ranges larger than what the robot may encounter in order to account for error accumulation over time. 

\begin{figure}
\centering
\subfloat[Background Textures]{\centering
\resizebox{0.495\columnwidth}{!}{\includegraphics{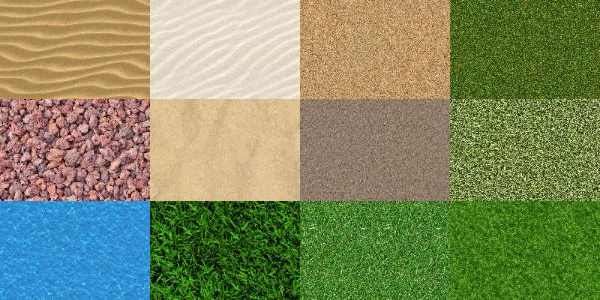}}
\label{fig:textures_back}}
\subfloat[Stone Textures]{\centering
\resizebox{0.495\columnwidth}{!}{\includegraphics{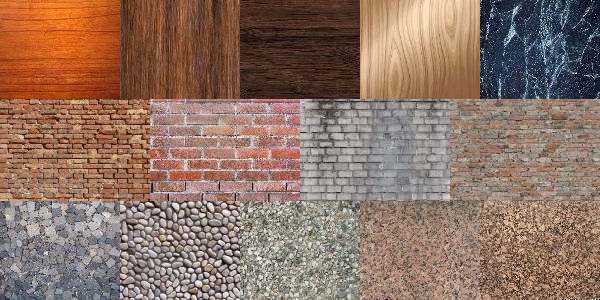}}
\label{fig:textures_stone}} 
\caption{A collage of textures use for the synthetic outdoor dataset generation: (a) Background Textures, (b) Stone Textures.}
\label{fig:textures}
\vspace{-15pt}
\end{figure}

\begin{figure}
\centering 
\includegraphics[width=0.7\columnwidth]{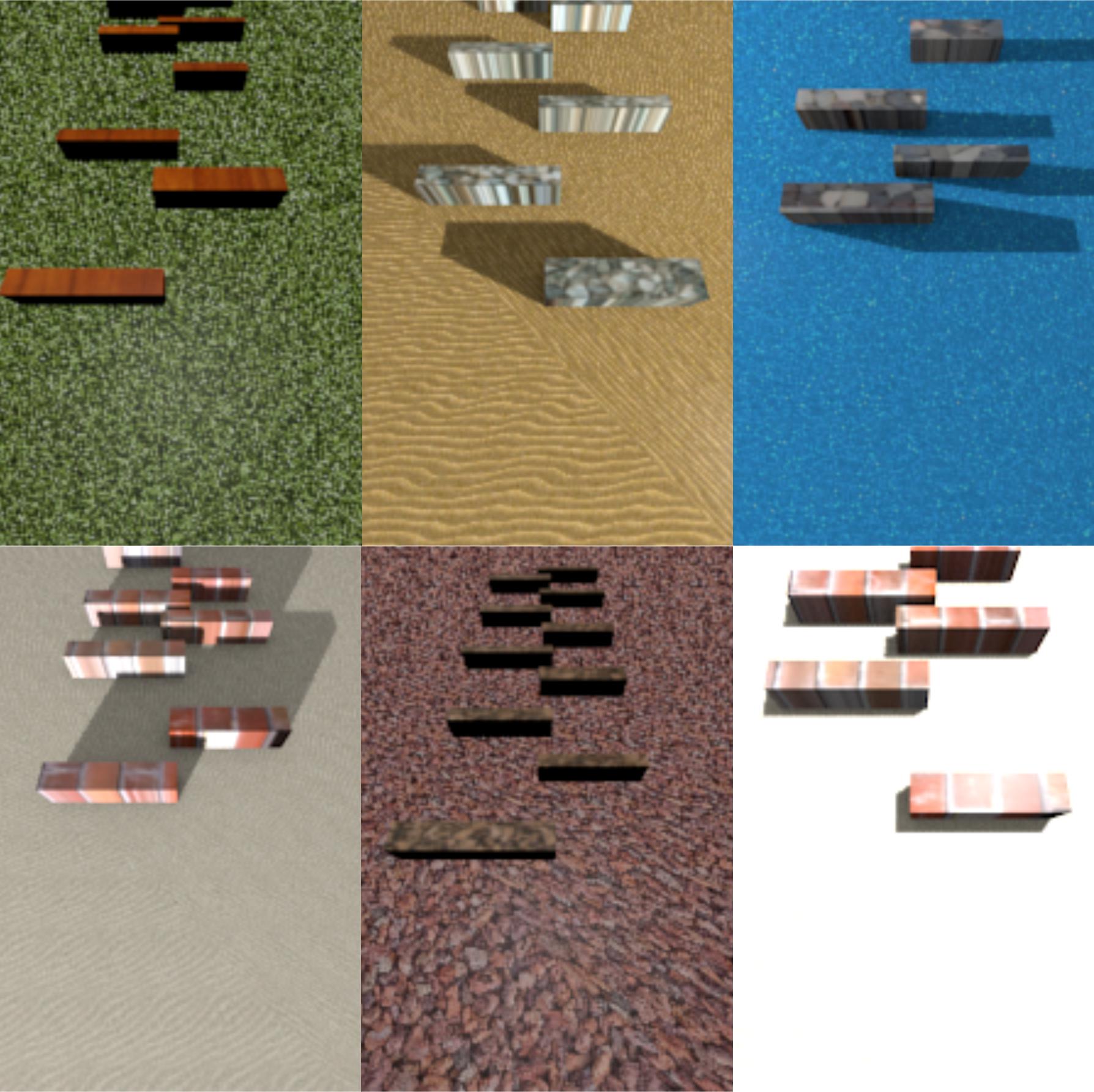}
\label{fig:sampleimages_sod} 
\caption{Sample images from the \emph{Synthetic Outdoor Dataset} (SOD).}
\label{fig:sampleimages}
\vspace{-15pt}
\end{figure}

\begin{figure*}
\centering
\includegraphics[width=0.7\textwidth]{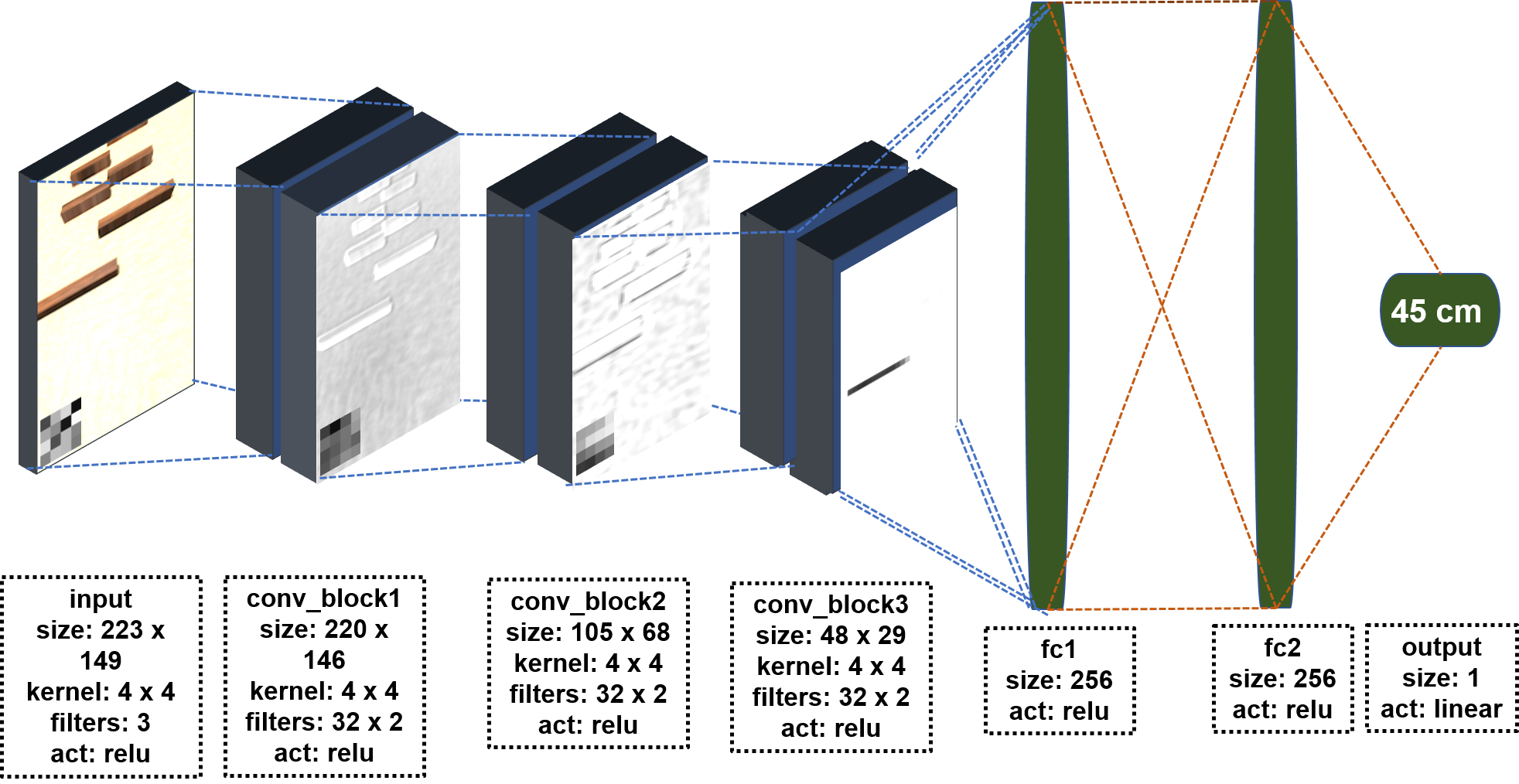}
\caption{Deep Network Architecture for predicting step length. Each convolutional block consists of two convolutional layers followed by a Max-Pool and Batch Normalization layers. Two identically dimensioned fully connected layers are used. Finally, the output activation function is linear.}
\label{fig:network}
\vspace{-10pt}
\end{figure*}	

The above four parameters are randomized for image generation in the following manner:

\textbf{Camera Location}: The camera location is measured with respect to the stance foot position. For each image, we randomly sampled from a range of $[-10:20]\times[-10:10]\times[80:120]$ (cm) to obtain the x, y, z offsets of the camera from the stance foot, respectively. The ZED Stereo Camera \cite{zed} model is used for rendering the images. However, since we only focus on the step length, we only generate monocular images for this study.

\textbf{Stone Location}: From \cite{RSS2017_DiscreteTerrain_Walking}, we note that, the robot was able to walk on a discrete terrain where the step lengths ranged from $[20:90]$ (cm). True step length is the distance from the robot's stance foot to the next stone's center. For this work we uniformly sample step lengths from $[15:95]$ (cm) range and arrange the stones in a single column accordingly. Based on the width of the robot, we spread out stones $[30:35]$ (cm) away of the robot's center-line. Moreover, they are alternately positioned on either sides of the center-line, as shown in Fig.~\ref{fig:sampleimages}. Finally, the stone size itself is varied randomly between $[10:20]$ (cm) in length and $[60:90]$ (cm) in width, respectively. However, the stone height is kept fixed at $15$ cm.

\textbf{Lighting Properties:} The light source is always facing the current stepping stone. However, its position is randomly chosen from a semi-spherical dome-like space above the robot with a $6$ meters constant radius to the current stone-center. 

\textbf{Texture:} Here we select real world textures for both stones and background. We chose $12$ unique textures comprising of grass, sand, water, pebbled terrain, etc. for the background and $14$ unique textures for the stone including granite, cement, brick, wood, etc. Texture samples are shown in Fig. \ref{fig:textures} Each scene is rendered by randomly sampling a texture pair from the available collection. 

The images are rendered with a resolution of $223 \times 223$ pixels. Additionally, we crop the left and right $15 \%$ of the image to further reduce computational overhead. The final image resolution is $223 \times 149$. We generate $40,000$ images and call it the Synthetic Outdoor Dataset (SOD). Sample images are shown in Fig. \ref{fig:sampleimages}. Having presented the dataset generation details, we will present the neural network architecture design next.

\subsection{Network Architecture Summary}
For training our object detector, we propose a custom neural network architecture. 
It consists of six convolutional layers of $32$ filters each, followed by two dense layers, both with $256$ neurons each. Unlike traditional designs where the number of features maps increase with depth, we found that, a constant number of feature maps throughout does better and needs fewer parameters. The kernel size of each filter is $(4\times 4)$. Batch Normalization and Max Pooling are applied after every two convolutional layers. Additionally, we apply Batch Normalization just before the final output layer as well. We use \emph{relu} activations in all the layers except the last one, where we use a \emph{linear} activation function instead. Fig.~\ref{fig:network} summarizes the CNN architecture. The detector is trained using the \emph{Mean Squared Error} loss and the \emph{Adam} optimizer. We use a learning rate of $1e-4$ along with a suitable learning rate decay policy. We use Keras API \cite{keras} with TensorFlow backend \cite{tensorflow} to build and train our model. We trained for $40$ epochs with a batch size of $50$, on a Intel i7 machine with an NVIDIA Titan X GPU. The model has roughly $2.5$ million trainable parameters. Finally we split the dataset into Training, Validation and Test sets, each comprising of $28900, 5100, 6000$ images, respectively.

The step length is estimated as follows: Using the joint encoders on the robot, the location of the camera is first estimated. The deep neural network only learns to predict the distance from the camera to the next stone center. Therefore, the predicted step length is the sum of these two distances. Note that the camera position information is not used during training. 

While deep networks have remarkable function approximation abilities, they have many tunable hyper-parameters whose fine-tuning critically impacts network performance and generalization ability. In the next section, we systematically outline our network design and customization process while examining the impact of each hyper-parameter on reducing worst-case test error. 
 
\subsection{Hyper-parameter Search}
Deep neural networks have a very high dimensional hyper-parameter space, where almost every single building block can be optimized. Most applied deep learning papers use existing architectures and leverage their transfer learning properties. Few papers explain in sufficient detail, the impact of each hyper-parameter on the learning outcomes. Unfortunately, hyper-parameter choices don't always generalize  to all problems and it is therefore worthwhile to carefully tweak and specifically examine them for individual problems. Important insights drawn from this exploration for our problem are summarized below. Note that, all the results reported below are on the test set.

\begin{itemize}
\item Roughly $89\%$ drop in error occurs within the first $20$ epochs. Therefore, a wider hyper-parameter coarse search was carried out on models trained for $20$ epochs while for the finer search, the models were trained for $40$ epochs. 
\item As already identified in \cite{tobin2017domain}, Dropout with any probability or placement in the network worsens performance. 
\item Batch Normalization really boosts performance and results in around $33\%$ drop in mean absolute error. More interestingly it leads to over $55\%$ drop in the worst-case prediction error\ (or the maximum prediction error on the test set). 
\item Using L2 Regularization (default is $0.01$) for only the fully connected layers helps further reduce the worst-case prediction error.
\item We tested the Architecture with 5 kernel sizes, $(2,2), (4,4), (6,6), (3,3), (2,4)$. We observed that rectangular kernels had the largest worst-case error, followed by the $(3,3)$ kernel. Surprisingly, even numbered kernels did a better job, against conventional wisdom. The best kernel was $(4,4)$.
\item Adding an extra convolutional block (ie., two additional convolutional layers followed by a max pool and a batch norm) increased the mean absolute error. Unlike classification tasks where depth almost always helps, in regression tasks, localization accuracy is affected by max-pooling layers beyond some depth. Therefore, for regression tasks one must find the sweet spot between depth (complexity) and accuracy.
\end{itemize}

Finally, based on the above mentioned hyper-parameter search, an optimal network architecture is designed and it is trained with the synthetic outdoor dataset. The qualitative and quantitative results obtained are presented in the next section.

\section{Results and Discussion}
\label{sec:results}
In this section, we study the performance of the deep perception model and then integrate it with a physics-based simulator and gait-library based controller to numerically realize and analyze autonomous dynamic walking on randomly generated discrete terrain.

\subsection{Step Length Prediction Performance}


Once trained, we expect the step length predictor to accurately detect the next stepping stone and output it's distance in centimeters. Note that, each image will have anywhere between $1-5$ stepping stones. Therefore, even though they have the same texture and geometry, our perception framework needs to overlook other stones and actively seek out the first one. In this situation, we believe perspective distortion helps in better distinguishing the stone of interest. Further, due to the safety-critical nature of this problem, in addition to describing network performance based on mean squared error, we will also report the variance and the worst-case prediction.  

\begin{figure}
\centering
\subfloat[Predicted v/s Actual Step Length ]{\centering
\resizebox{0.48\columnwidth}{!}{\includegraphics{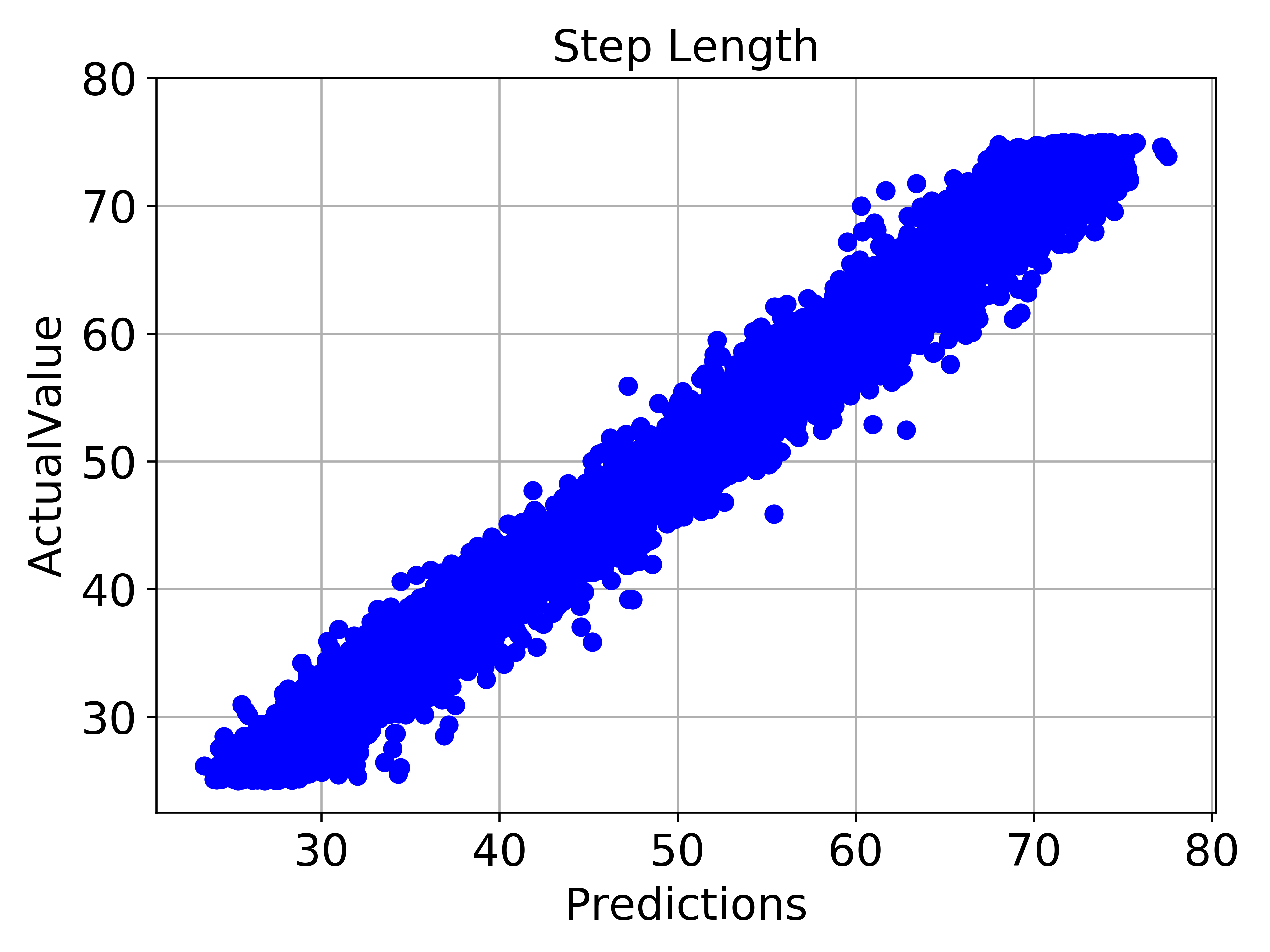}}
\label{fig:predvsact}} 
\subfloat[Prediction Error Histogram]{\centering
\resizebox{0.48\columnwidth}{!}{\includegraphics{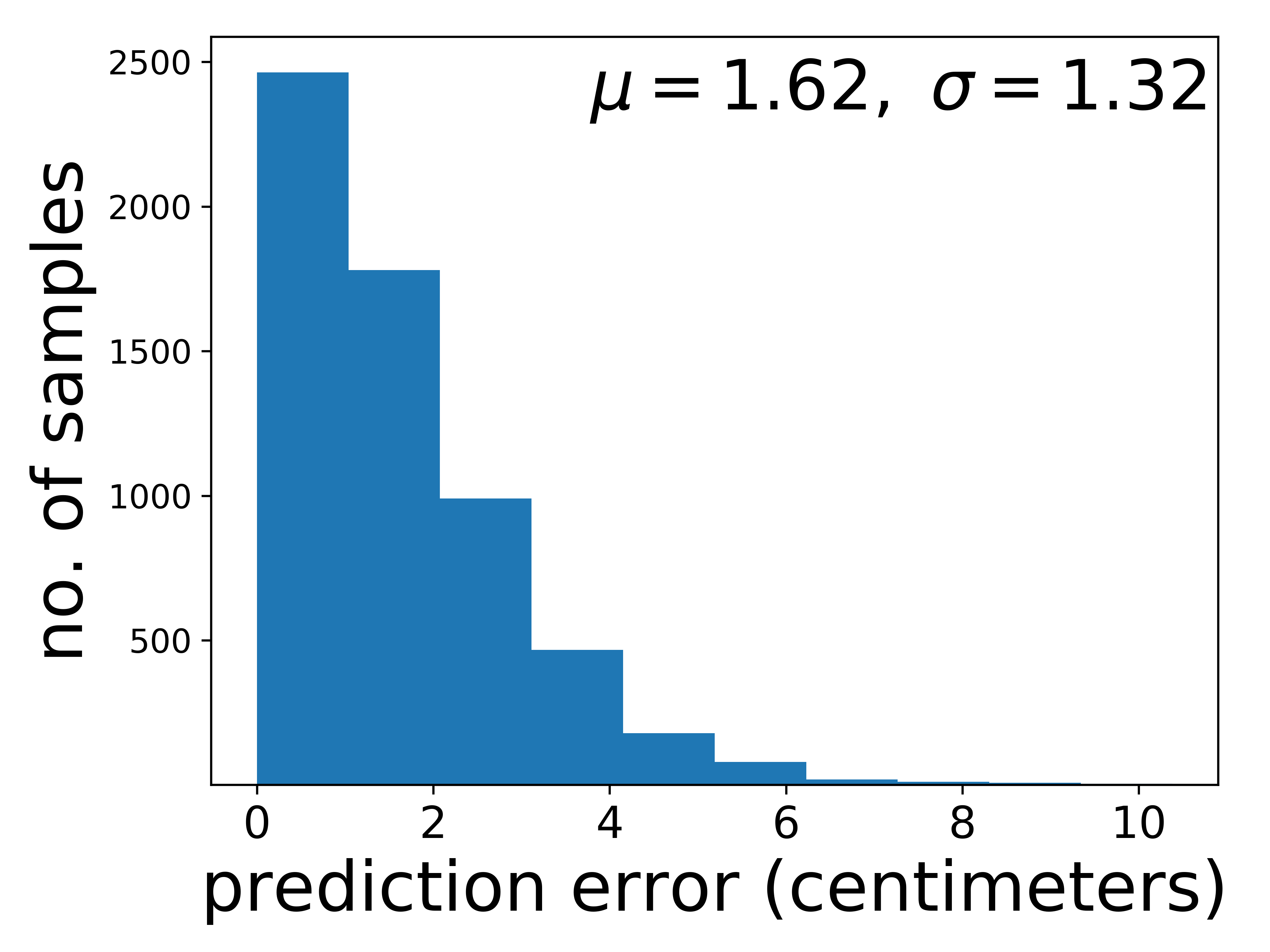}}
\label{fig:errhist}} 
\caption{Visualizing step length prediction performance through plots of (a) Predicted versus True Step Length values and (b) Prediction Error Histogram. Both plots are for the test data obtained from the Synthetic Outdoor Dataset.}
\label{fig:steplenpred}
\vspace{-15pt}
\end{figure}

Error is unavoidable in function approximation. However, in a safety-critical scenario like discrete terrain walking, the default approach of choosing the network that gives the least mean squared error could be detrimental as the worst-case estimate could still be off the safety limits. In order to avoid this issue, in this work, hyper-parameters were tuned with the objective to find the least possible worst-case estimate with the available dataset. The performance of the CNN was evaluated on the $6000$ sample test data and is visualized in Fig.~\ref{fig:steplenpred} and summarized in Table \ref{table:trainingperformance}.

\begin{table}
\resizebox{\columnwidth}{!}{
\centering
\begin{tabular}{cccc}
\toprule
 Test Avg. Loss & Std. Dev. of Loss  & \% Above 5 cm & Max Pred. Error \\ 
 \quad (in cm) &   &  & \quad (in cm)  \\ \midrule
 1.618 & 1.32 & 2.116 & 10.38 \\ \bottomrule
\end{tabular}}
\caption{Summary of prediction performance on test data.}
\label{table:trainingperformance}
\vspace{-20pt}
\end{table}

\subsubsection{Best and Worst predictions:} 
\begin{figure}
	\centering
	\subfloat[Best-8]{\centering
		\resizebox{0.85\columnwidth}{!}{\includegraphics{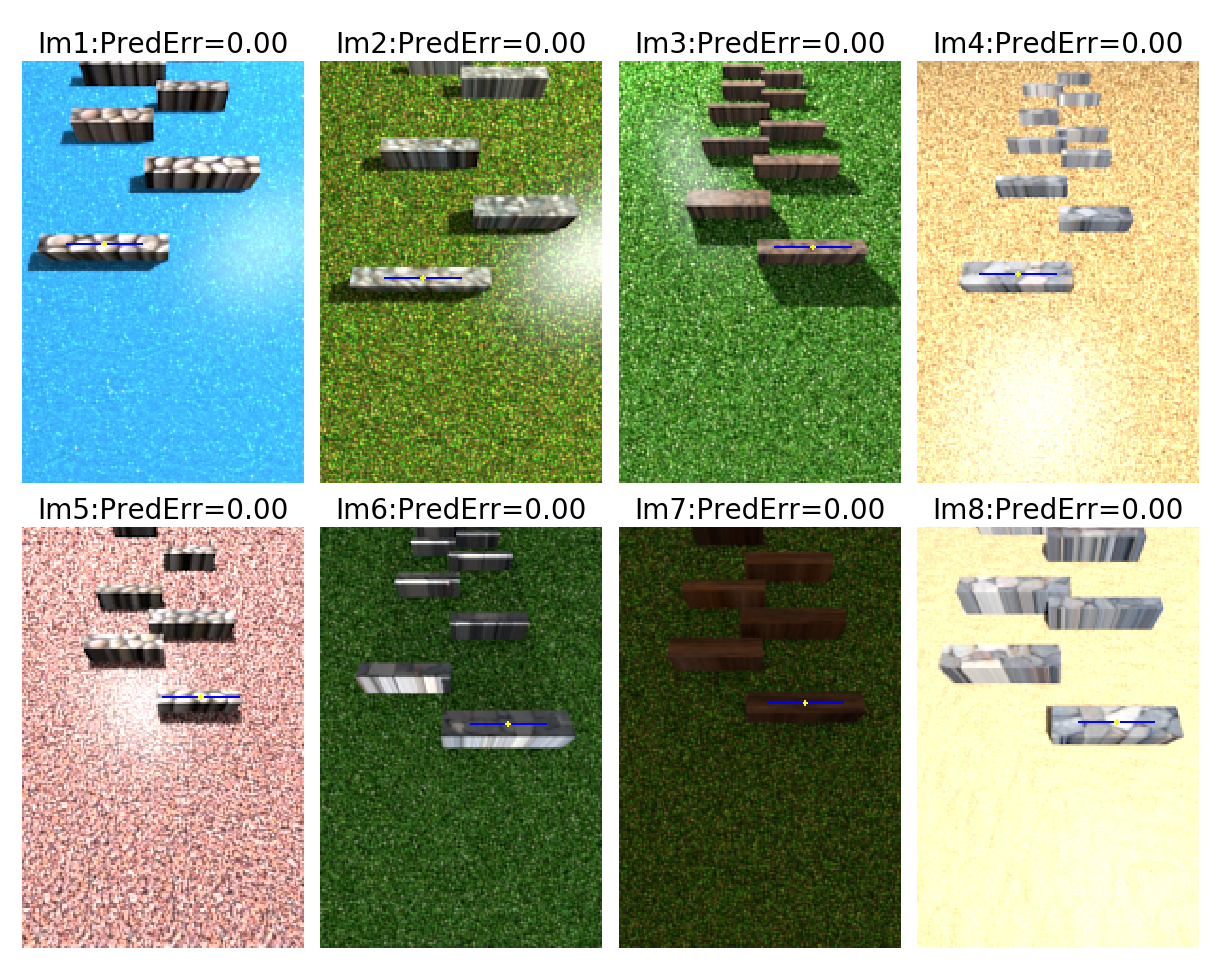}}
\label{fig:bestn}} \\%
	\vspace{-10pt}
	\subfloat[Worst-8]{\centering
		\resizebox{0.85\columnwidth}{!}{\includegraphics{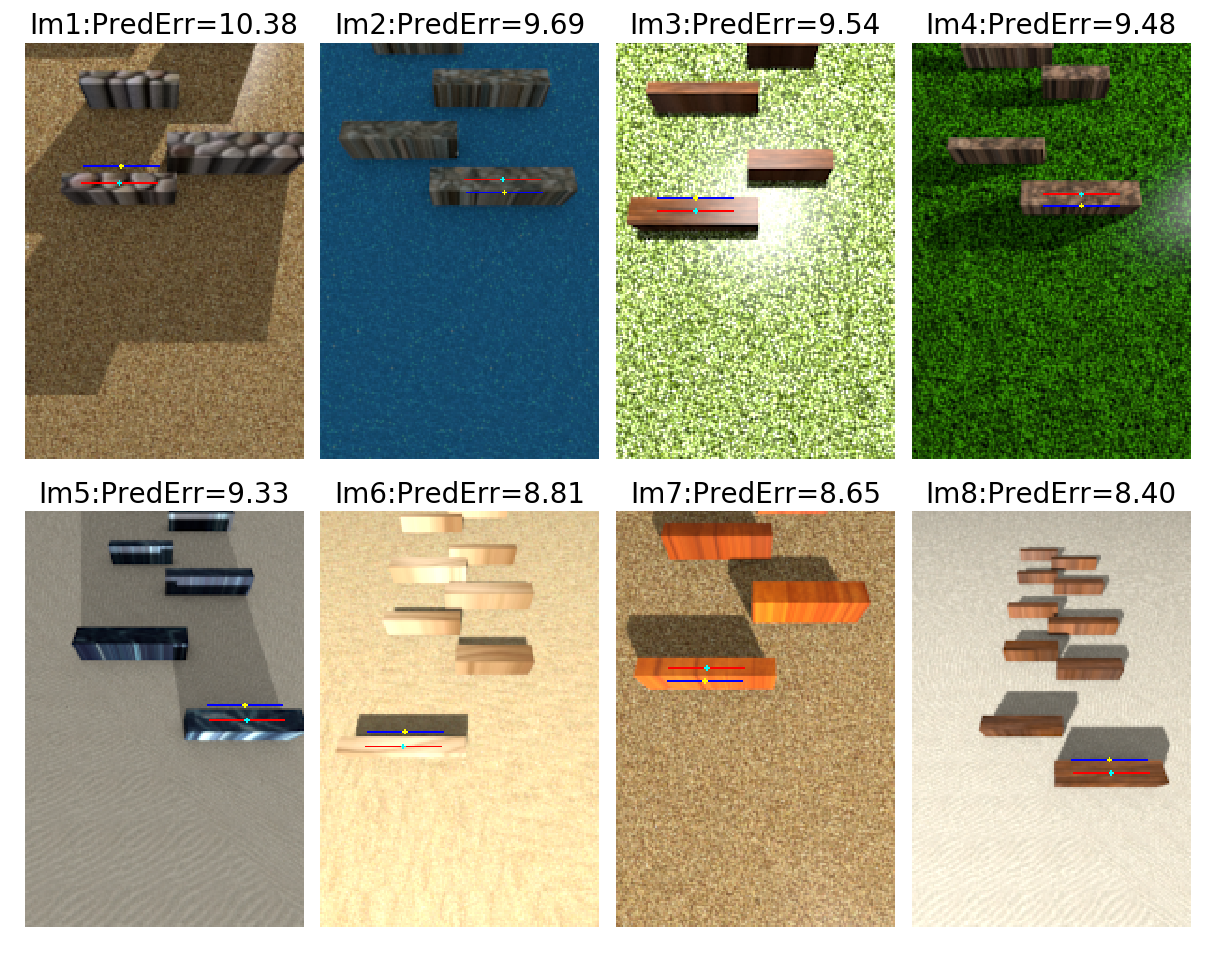}}
\label{fig:worstn}}
	\caption{Snapshots of the best-8 and worst-8 predictions of the neural network alongwith the corresponding prediction error in centimeters. The desired (red line) and predicted (blue line) step lengths are marked along with pixel coordinates of the resulting foot placement location (yellow dot).}
	\label{fig:bestnworstn}
    \vspace{-15pt}
\end{figure}

In addition to studying the qualitative learning outcomes like average loss, standard deviation of loss, etc., we also visualize images of the best-8 and the worst-8 predictions in Fig. \ref{fig:bestnworstn} to visually interpret which parameters the model could and couldn't generalize over. From the figure, it is clear that the model is able to generalize over the various foreground and background textures, including bright and dim lighting conditions. However, shadows contributed to some of the higher estimation errors. 

\subsection{Simulation Results}

\begin{figure*}
\centering			             
\includegraphics[width=0.7\textwidth]{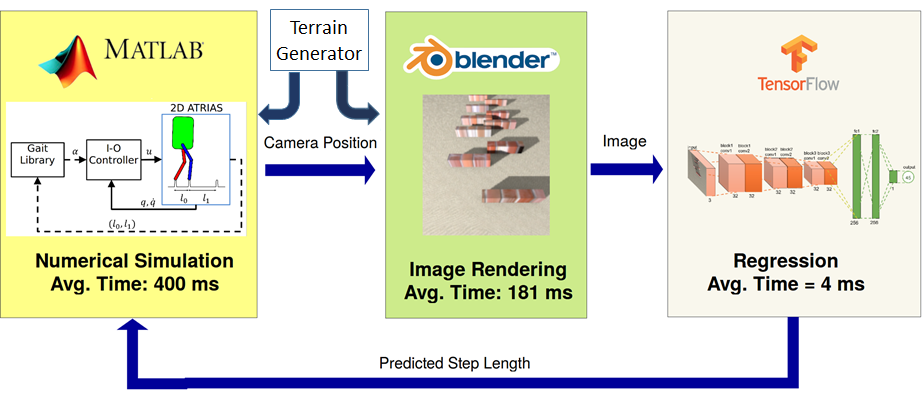}
\vspace{-10pt}
\caption{Simulation pipeline for autonomous dynamic walking on discrete terrain. The pipleline integrates gait optimization, nonlinear control, vision, and deep learning. Simulation Video: \url{https://youtu.be/ijJAPapU7qI}.}
\label{fig:sim_schematic}
\vspace{-15pt}
\end{figure*}

\begin{figure}
	\centering
	\subfloat[Prediction Error]{\centering
		\resizebox{0.8\columnwidth}{!}{\includegraphics{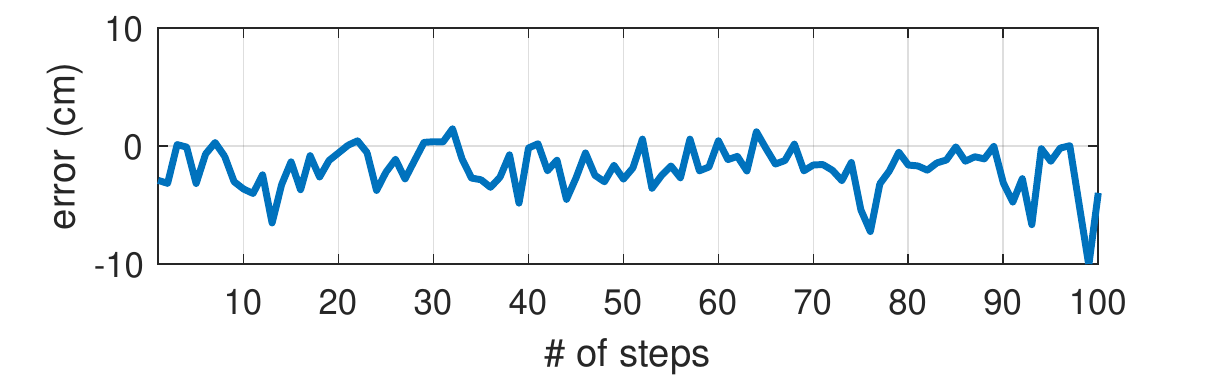}}
\label{fig:pe_50to80}} \\
	\subfloat[Foot Placement Error]{\centering
		\resizebox{0.8\columnwidth}{!}{\includegraphics{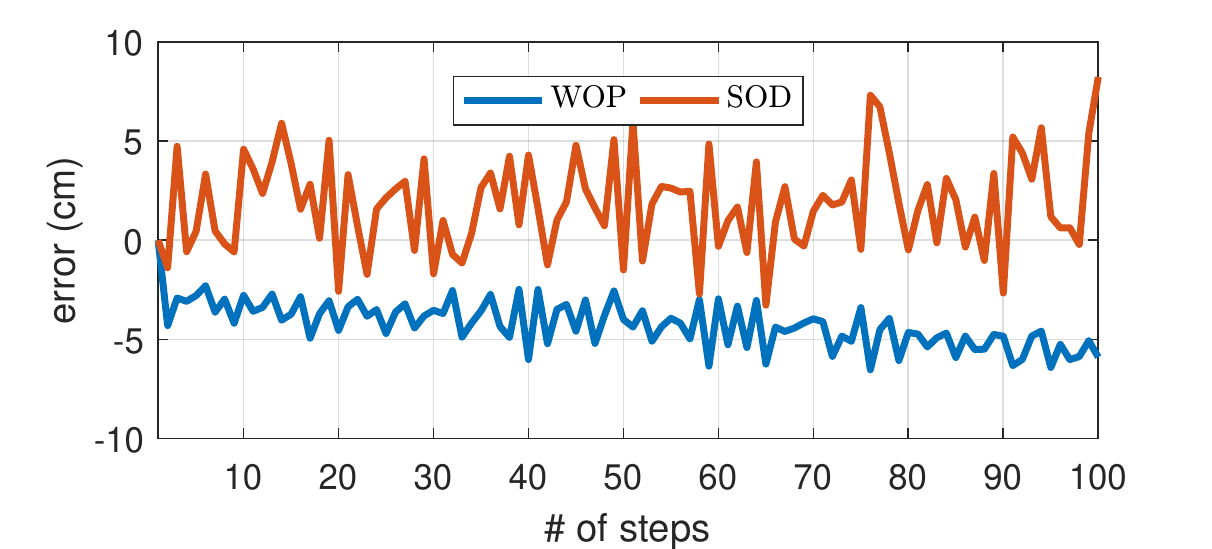}}
\label{fig:fpe_50to80}}
	\caption{(a) Prediction Error, and (b) Foot Placement Error plots for 100 step walking simulation with step lengths varying within $[45:75]$ cm. In (b) WOP indicates error \emph{without perception} while SOD indicates errors when step length is estimated using the \emph{synthetic outdoor dataset}-based image preview. }
	\label{fig:errors50to80}
\vspace{-15pt}
\end{figure}

In this section, we integrate the CNN estimator with a physics-based robot simulator in \ref{subsec:model} and the gait-library based controller in \ref{subsec:gait_library} to evaluate the closed-loop autonomous operation of the robot. The robot dynamics are simulated in Matlab. At the beginning of each step, robot's current position (specifically the camera position) is supplied to Blender to render a first-person-view synthetic outdoor image using the information from the terrain generator. This image is in turn supplied to the Convolution Neural Network (implemented in Keras and TensorFlow) to predict the step length for the next step. Provided with this predicted step length, the closed-loop dynamics of the robot and controller are simlated for one step to enable the robot to take the step forward. The process is repeated for subsequent steps. A schematic of the numerical simulation pipeline is shown in Fig. \ref{fig:sim_schematic}.

Numerical simulations are carried out with stones randomly placed with the inter-stone distance within the $[45 : 85]$ cm range. Recall that the camera position and stone location were uniformly sampled from dissimilar ranges in Section IV. The label used for training the CNN is the distance to the camera which is difference of stone location and camera position. Therefore, the distribution of labels used for training is no longer uniform. Accounting for this fact, the step length ranges have been adjusted to only test the CNN in the range where there was enough data to guarantee a good learning outcome.

In this study, we render images with light fixed in an overhead position and focus on carefully examining sensitivity to  potential failure modes like camera position or step length going outside the range used for dataset generation during continuous simulation.  (We have already noted earlier that shadows are a failure mode.) Our evaluation is based on two metrics, 1) Perception Error 
and 2) Foot Placement Error. 
Prediction Error is defined as the difference between Predicted Step Length and True Step Length and it is purely an artifact of the perception module. Similarly, Foot Placement Error is defined as the difference between foot-contact-point and stone-center. This error is the cumulative error of both perception and control. 

The robot was simulated for $100$ steps and these two errors are plotted, as shown in Fig. \ref{fig:errors50to80}. Note that the prediction error is bounded within a $5$ cm range from the center for the most part and doesn't show significant accumulation of error over time. In Fig. \ref{fig:errors50to80}(b), the foot placement error is also plotted for the case where perception was not used (called WOP - With Out Perception) and with perception (using SOD - synthetic outdor dataset) for comparison. In an attempt to minimize effort, the conservative controller always under-steps. In contrast, the prediction model over-estimates more frequently. Therefore, the errors cancel more often and lead to a desirable and mostly-bounded stepping behavior.  

\section{Conclusions and Future Work}
\label{sec:conc}
In this work, we outlined a systematic way to design a CNN-based predictor that can estimate step lengths for a dynamic bipedal walker operating in discrete terrain. It is shown empirically that a feed-forward gait adjustment based on intermittent visual feedback is sufficient to walk on a discrete terrain where high-speed prediction and accurate foot placement are critical. Several visual factors that impact the predictor's performance are identified for further refinement.
This paper integrates gait optimization, nonlinear control, vision, and deep learning.

As part of future work, the direct perception model will be extended to predict other related gait parameters like step width, step height and yaw angle, enabling autonomous rough terrain navigation. Moreover, since multiple steps are visible in an image, accuracy could be improved through Recurrent Neural Networks. 
Domain adaptation techniques that transfer the predictor from simulation to real world images will enable testing the algorithm on the real robot.

\balance
\bibliographystyle{IEEEtranS}
\bibliography{newrefs,ref,Quan_ref} 

\end{document}